\documentclass{article}

\usepackage[final]{nips_2017}

\usepackage{array}
\newcolumntype{x}[1]{>{\centering\let\newline\\\arraybackslash\hspace{0pt}}p{#1}}

\usepackage{placeins}

\usepackage[utf8]{inputenc} 
\usepackage[T1]{fontenc}    
\usepackage{hyperref}       
\usepackage{url}            
\usepackage{booktabs}       
\usepackage{amsfonts}       
\usepackage{nicefrac}       
\usepackage{microtype}      
\usepackage{graphicx}
\usepackage{amsmath,amsfonts,amsthm,bm} 
\usepackage{mathtools}
\usepackage{resizegather}
\usepackage{tikz}
\usepackage{graphbox}
\usepackage{subcaption}
\usepackage{array,booktabs}
\usetikzlibrary{bayesnet}
\graphicspath{{figures/}}
\title{Nonparametric Inference for Auto-Encoding Variational Bayes}

\author{
  Erik Bodin~\textsuperscript{*}\\
\And
  Iman Malik~\textsuperscript{*}\\
\And
  Carl Henrik Ek~\textsuperscript{*}\\
 \And
  Neill D. F. Campbell~\textsuperscript{$\dagger$}\\
\AND
\vspace{-0.5cm}\textsuperscript{*}~University of Bristol $\quad$ \textsuperscript{$\dagger$}~University of Bath
}

\begin{document}

\maketitle



Variational approximations are an attractive approach for inference of latent variables in unsupervised learning. However, they
are often computationally intractable when faced with large datasets. Recently, Variational Autoencoders (VAEs) \cite{Kingma2013} have been proposed as a method to tackle this limitation. Their methodology is based on formulating the approximating posterior distributions in terms of a deterministic relationship to observed data consequently the title ``Auto-Encoding Variational Bayes''. Importantly, this is a decision regarding an approximate inference scheme that should not be confused with an auto-encoder as a model.

Unsupervised learning is a ill-conditioned problem that requires prior knowledge to reach a solution. We would like to learn latent representations that are low-dimensional and highly interpretable. A model that has these characteristics is the Gaussian Process Latent Variable Model (GP-LVM) \cite{Lawrence:2005vk}. The benefits and negative of the GP-LVM are complementary to the VAE, the former provides useful low-dimensional latent representations while the latter is able to handle large amounts of data and can use non-Gaussian likelihoods. Our inspiration for this paper is to marry these two approaches and reap the benefits of both. In order to do so we will introduce a novel approximate inference scheme inspired by the GP-LVM to the VAE. 
%
%

The standard VAE formulation \cite{Kingma2013} adopts a unit Gaussian prior, creating a trade-off between the embedded data residing at the same location in the latent space and the ability to reconstruct the data in the observed space. This encourages a ``tight packing'' of the data around a shared origin, with hope that similar data in the observed space have overlapping probability mass in the latent space i.e. mutual information. It has been shown that a simple prior over-regularizes the latent space leading to poor reconstructions \cite{hoffman2016elbo}. Other more flexible priors can be used to change the dynamic between reconstruction and mutual information, such as in recent work of using a mixture \cite{tomczak2017vae} or let the prior be autoregressive \cite{chen2016variational}. The quality of the output from a VAE can be improved by more expressive generative models, but this has been shown to lead to a tendency of ignoring the latent space, defeating the purpose of unsupervised learning \cite{zhao2017towards}. In summary, reconstruction quality and mutual information in the latent space is traded against each other.

In this paper we address this limiting trade-off by escaping it; we let the space where we encourage sharing be separated from the space where the generative capacity is set. We do this by an approximation of a model where the observations can be generated from either space. We show experimentally that the approximation allows the capacity of the generative bottle-neck ($Z$) to be arbitrarily large without losing sharing and the beneficial properties of the sharing space ($X$), allowing reconstruction quality to be unlimited by $Z$ at the same time as a low-dimensional space can be used to perform ancestral sampling from as well as a means to reason about the embedded data.

\section{Method}

\begin{figure}[t!]
\centering
\hspace*{0.2in}
  \begin{tikzpicture}

  \node[obs]                 (y) {$\mathbf{Y}$};
  \node[latent, above=of y]  (z) {$\mathbf{Z}$};

  \path (z) -- node[auto=false, yshift=-1.0cm, xshift=1.2cm]{$\rightarrow$} (z);

  \edge {z} {y} ;

  \end{tikzpicture}
\hspace*{0.1in}
  \begin{tikzpicture}

  \node[obs]                 (y1) {$\mathbf{y}_{1}$};
  \node[latent, above=of y1]  (z1) {$\mathbf{z}_{1}$};

  \node[obs, right=0.3cm of y1]      (y2) {$\mathbf{y}_{2}$};
  \node[latent, above=of y2]   (z2) {$\mathbf{z}_{2}$};

  \node[obs, right=0.8cm of y2]     (yn) {$\mathbf{y}_{N}$};
  \node[latent, above=of yn]  (zn) {$\mathbf{z}_{N}$};

  \path (zn) -- node[auto=false, yshift=-1.0cm, xshift=1.2cm]{$\rightarrow$} (zn);
  \path (zn) -- node[auto=false, xshift=-0.75cm]{\ldots} (yn);

  \edge {z1} {y1} ;
  \edge {z2} {y2} ;
  \edge {zn} {yn} ;

  \end{tikzpicture}
\hspace*{0.1in}
  \begin{tikzpicture}

    \node[obs]                 (y1) {$\mathbf{y}_{1}$};
    \node[latent, above=of y1]  (z1) {$\mathbf{z}_{1}$};

    \node[obs, right=1cm of y1]  (y2) {$\mathbf{y}_{2}$};
    \node[latent, above=of y2]  (z2) {$\mathbf{z}_{2}$};

    \node[obs, right=4cm of y1] (yn) {$\mathbf{y}_{N}$};
    \node[latent, above=of yn]  (zn) {$\mathbf{z}_{N}$};

    \node[det, above=of z2, yshift=-0.5cm, xshift=0.5cm, fill=gray!25]  (Y) {$\mathbf{Y}$};

    \path (zn) -- node[auto=false, xshift=-1.5cm]{\ldots} (yn);

    \edge {Y} {z1} ;
    \edge [dash pattern=on 6pt off 10pt on 10pt] {z1} {y1} ;

    \edge {Y} {z2} ;
    \edge [dash pattern=on 6pt off 10pt on 10pt] {z2} {y2} ;

    \edge {Y} {zn} ;
    \edge [dash pattern=on 6pt off 10pt on 10pt] {zn} {yn} ;

  \end{tikzpicture}\\[10pt]
\setlength{\tabcolsep}{0pt}
\begin{tabular}{x{0.2\textwidth}x{0.3\textwidth}x{0.5\textwidth}}
\textbf{(a)}~Model & \textbf{(b)}~VAE~Approximation & \textbf{(c)}~VAE~Inference~Scheme
\end{tabular}\\[16pt]
%
%
%
  \begin{tikzpicture}[baseline={(0,-2.8)}]

  \node[latent]     [yshift=0.5cm]            (y) {};
  \node[below=of y, xshift=0cm, yshift=1.0cm] (yt) {\small{Distribution}};
  \node[det, below=of y, yshift=0.4cm]  (z) {};
  \node[below=of z, xshift=0cm, yshift=1.0cm] (zt) {\small{Point Estimate}};
  
  \node[latent, rectangle, draw=white, fill=gray!25, below=of z, yshift=0.4cm] (a) {};
  \node[below=of a, xshift=0cm, yshift=1.0cm] (at) {\small{Observed}};

  \end{tikzpicture}
    \hspace*{0.2in}
  \begin{tikzpicture}
    \node[obs]                 (y1) {$\mathbf{y}_{1}$};
    \node[latent, above=of y1]  (z1) {$\mathbf{z}_{1}$};

    \node[obs, right=0.3cm of y1]  (y2) {$\mathbf{y}_{2}$};
    \node[latent, above=of y2]  (z2) {$\mathbf{z}_{2}$};

    \node[obs, right=0.8cm of y2] (yn) {$\mathbf{y}_{N}$};
    \node[latent, above=of yn]  (zn) {$\mathbf{z}_{N}$};

    \node[latent, above=of z2, yshift=-0.5cm, xshift=0.5cm]  (X) {$\mathbf{X}$};

    \path (zn) -- node[auto=false, xshift=-0.75cm]{\ldots} (yn);
    \path (zn) -- node[auto=false, yshift=-0.75cm, xshift=1.2cm]{$\rightarrow$} (zn);

    \edge {X} {z1} ;
    \edge {z1} {y1} ;

    \edge {X} {z2} ;
    \edge {z2} {y2} ;

    \edge {X} {zn} ;
    \edge {zn} {yn} ;
  \end{tikzpicture}
\hspace*{0.1in}
  \begin{tikzpicture}
    \node[obs]                 (y1) {$\mathbf{y}_{1}$};
    \node[latent, above=of y1]  (z1) {$\mathbf{z}_{1}$};

    \node[obs, right=0.3cm of y1]  (y2) {$\mathbf{y}_{2}$};
    \node[latent, above=of y2]  (z2) {$\mathbf{z}_{2}$};

    \node[obs, right=0.8cm of y2] (yn) {$\mathbf{y}_{N}$};
    \node[latent, above=of yn]  (zn) {$\mathbf{z}_{N}$};

    \node[det, above=of zn, yshift=-0.25cm, xshift=-1.25cm, fill=gray!25]  (Y) {$\mathbf{Y}$};
    \node[det, right=of Y, xshift=1cm]  (X) {$\mathbf{X}$};
    
      \node[latent, below=of X, yshift=0.25cm]  (Z) {$\mathbf{\tilde Z}$};

    \path (zn) -- node[auto=false, xshift=-0.75cm]{\ldots} (yn);

    \edge {X} {Z};

    \edge [dash pattern=on 6pt off 10pt on 10pt] {z1} {y1} ;

    \edge [dash pattern=on 6pt off 10pt on 10pt] {z2} {y2} ;

    \edge [dash pattern=on 6pt off 10pt on 10pt] {zn} {yn} ;
    
    \edge {Y} {z1} ;
    \edge {Y} {z2} ;
    \edge {Y} {zn} ;

    \edge {Y} {X} ;
  \end{tikzpicture}\\[10pt]
\setlength{\tabcolsep}{0pt}
\begin{tabular}{x{0.2\textwidth}x{0.3\textwidth}x{0.5\textwidth}}
 & \textbf{(d)}~Our Nonparametric VAE & \textbf{(e)}~Our~Inference~Scheme
\end{tabular}
\caption{Contrasting variational approximation schemes for unsupervised learning. (a)~We specify an unsupervised generative model from latent $Z$ to observed $Y$. (b)~The VAE proposes a fully factored variational approximation. (c)~Inference proceeds by conditioning the variational latent parameters on observed data through an explicit deterministic function (e.g.~MLP network). (d)~Our model proposes an additional latent space $X$ that ties together the factored $Z$ space. (e)~Inference then proceeds with $X$ also conditioned on observed data through an additional deterministic function. For tractable inference we match moments between $Z$ and $\tilde Z$. }
\label{fig:graphical_models}
\vspace{-0.5cm}
\end{figure}
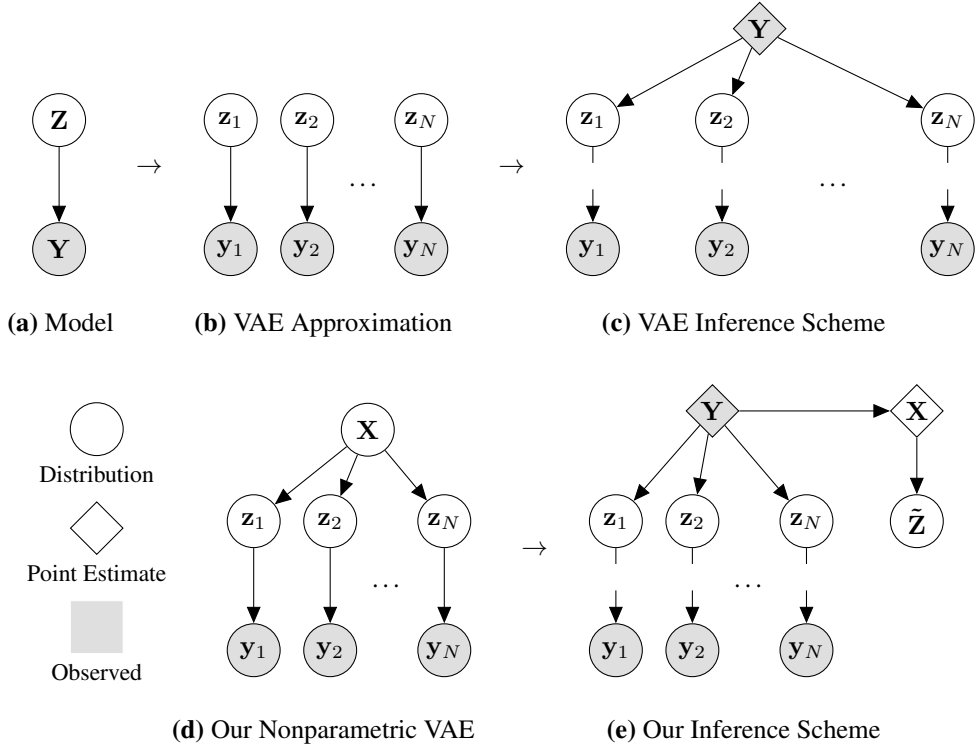

The VAE \cite{Kingma2013} inference scheme optimises a traditional evidence lower bound where the latent space posterior is approximated as a deterministic relationship from the observed data as $q(\mathbf{Z}|\mathbf{Y}) = \prod_{i=1}^Nq(\mathbf{z}_i|\mathbf{y}_i)$ where each latent variable is conditionally independent given the observed data. In this paper we introduce an additional latent variable $\mathbf{X}$ that model the interaction between the latent variables $\mathbf{Z}$.

Our approach means that $\mathbf{Z}$ are no longer independent but conditionally independent given $\mathbf{X}$. This leads to the following updated evidence lower bound with an additional divergence term as,
\begin{gather}\label{eq:lowerbound_w_extension}
\begin{split}
\mathcal{L}_{s} = \mathcal{L}_{g}- \text{KL}(q_{g}(\mathbf{Z}|\mathbf{Y})\vert\vert p_{s}(\mathbf{Z}|\mathbf{X})),
\end{split}
\end{gather}
where $\mathcal{L}_{g}$ and $q_{g}(\mathbf{z}_i|\mathbf{y}_i)$ is the standard VAE lower bound and approximative posterior respectively. To facilitate the use of batch processing we will rather than matching the joint distribution of the latent space match the predictive posteriors. As each $\mathbf{Z}$ is conditionally independent given the observed data $\mathbf{Y}$ this leads to the following updated objective function,
\begin{gather}\label{eq:lowerbound_w_extension2}
\begin{split}
\mathcal{L}'_{s} = \mathcal{L}_{g}- \text{KL}(\prod_{i}q_{g}(\mathbf{z}_i|\mathbf{y}_i)\vert\vert \prod_{i}p_{s}(\mathbf{z}_i|\mathbf{X},\mathbf{z}_{\neg i})).
\end{split}
\end{gather}
The predictive posterior of the Gaussian process is,
\begin{align}
  p(\mathbf{z}_i|\mathbf{X},\mathbf{z}_{\neg i}) = \mathcal{N}(&k(\mathbf{x}_i,\mathbf{X}_{\neg i})k(\mathbf{X}_{\neg i},\mathbf{X}_{\neg i})^{-1}\mathbf{z}_{\neg i},\nonumber\\
  &k(\mathbf{x}_i,\mathbf{x}_i)-k(\mathbf{x}_i,\mathbf{X}_{\neg i})k(\mathbf{X}_{\neg i},\mathbf{X}_{\neg i})k(\mathbf{X}_{\neg i},\mathbf{x}_i))
\end{align}
where $k(\cdot,\cdot)$ is the covariance function. Evaluating this posterior for $N$ datapoints is computationally expensive due to the inverse of the covariance function. To proceed we will introduce two additional approximations, first we will approximate the mean of the predictive posterior of the GP by directly parametrising it as,
\begin{align}
  \mu(\mathbf{z}_i) = k(\mathbf{x}_i,\mathbf{X}_{\neg i})k(\mathbf{X}_{\neg i},\mathbf{X}_{\neg i})^{-1}\mathbf{z}_{\neg i} \approx \mathbf{W}_{i}\mathbf{z}_{\neg i}.
\end{align}
In specific we will parametrise the weight matrix $\mathbf{W}_{i} \in \mathbb{R}^{1 \times N}$ such that the approximative predictive mean for $\mathbf{z}_i$ is a convex combination of $\mathbf{z}_{\neg i}$, i.e. that $\sum_{j} W_{ij} = 1, \forall {i}$ and $W_{ij} \geq 0, \forall {i,j}$. The intuition behind this is that we want to encourage sharing in the latent space $X$ such that the space $Z$ is represented in a distributed fashion. Secondly rather than minimising the KL-divergence we will match the first mode of the two distributions. This leads to our final objective function,
\begin{gather}
\begin{split}
\tilde{\mathcal{L}}_{s} = \mathcal{L}_{g}- \sum_{i} (\mathbb{E}[\mathbf{z}_{i}] - \mathbb{E}[\mathbf{\tilde z}_{i}])^{2},
\end{split}
\end{gather}
where $\mathbf{\tilde z}$ is the prediction of the latent space from $X$ while $\mathbf{z}$ is the prediction using the approximative posterior from the VAE. In effect we have separated the two models while retaining a connection by matching their first modes when predicting the latent space $Z$. We will now proceed to show the experimental evaluation of the model showing that we are capable of using the additional low-dimensional latent space $X$ as a proxy for the VAE latent space $Z$.

\section{Experiments}

In this paper we choose to model $W_{i, j}$ as a squared exponential covariance kernel between input locations, $\mathbf{x}_i$ and $\mathbf{x}_j$,
\begin{gather}
\begin{split}
W_{i, j} = k(\mathbf{x}_{i}, \mathbf{x}_{j}) = \sigma^{2} \exp\left({-\frac{\|\mathbf{x}_{i} - \mathbf{x}_{j}\|^{2}}{2l^{2}}}\right) \qquad i \neq j
\end{split}
\end{gather}
and infer its parameters.
We ensure $W_{i, i} = 0$ by subtracting the diagonal from the computed $\mathbf{W}$ and normalise by rows to create a convex combination. The latent locations $X$ are represented implicitly as a function of the observed data as in~\cite{Lawrence:2006wr}.

We trained models to illustrate that our extension can be used to obtain a low-dimensional $X$ space that is highly interpretable while permitting the use of a high-dimensional $Z$ space to provide high quality data generation. We validate our approach, with comparison to a standard VAE, by showing data embeddings in the $X$ space and generation of new data.

All experiments were performed with the decoder and both encoders as Multilayer Perceptrons (MLP) with the same architecture as in the original VAE \cite{Kingma2013}; we used two hidden layers of $500$ units each, mini-batch sizes of $128$ and a drop-out probability of $0.9$ throughout training. The decoder used was the Bernoulli MLP variant. Furthermore, the ADAM \cite{kingma2014adam} optimiser was used with a learning rate of $10^{-3}$. We varied the dimensionality of the inner most layer of the autoencoder (the $Z$ space) for the experiments. We used the MNIST data set from \cite{LeCun1998} comprised of $55~000$ training examples and $10~000$ test examples of $28 \times 28$ pixel greyscale images, corresponding to $784$ data dimensions.

\begin{figure}[h]
  \begin{minipage}{\textwidth}
    \centering
    \includegraphics[align=c,width=0.33\textwidth]{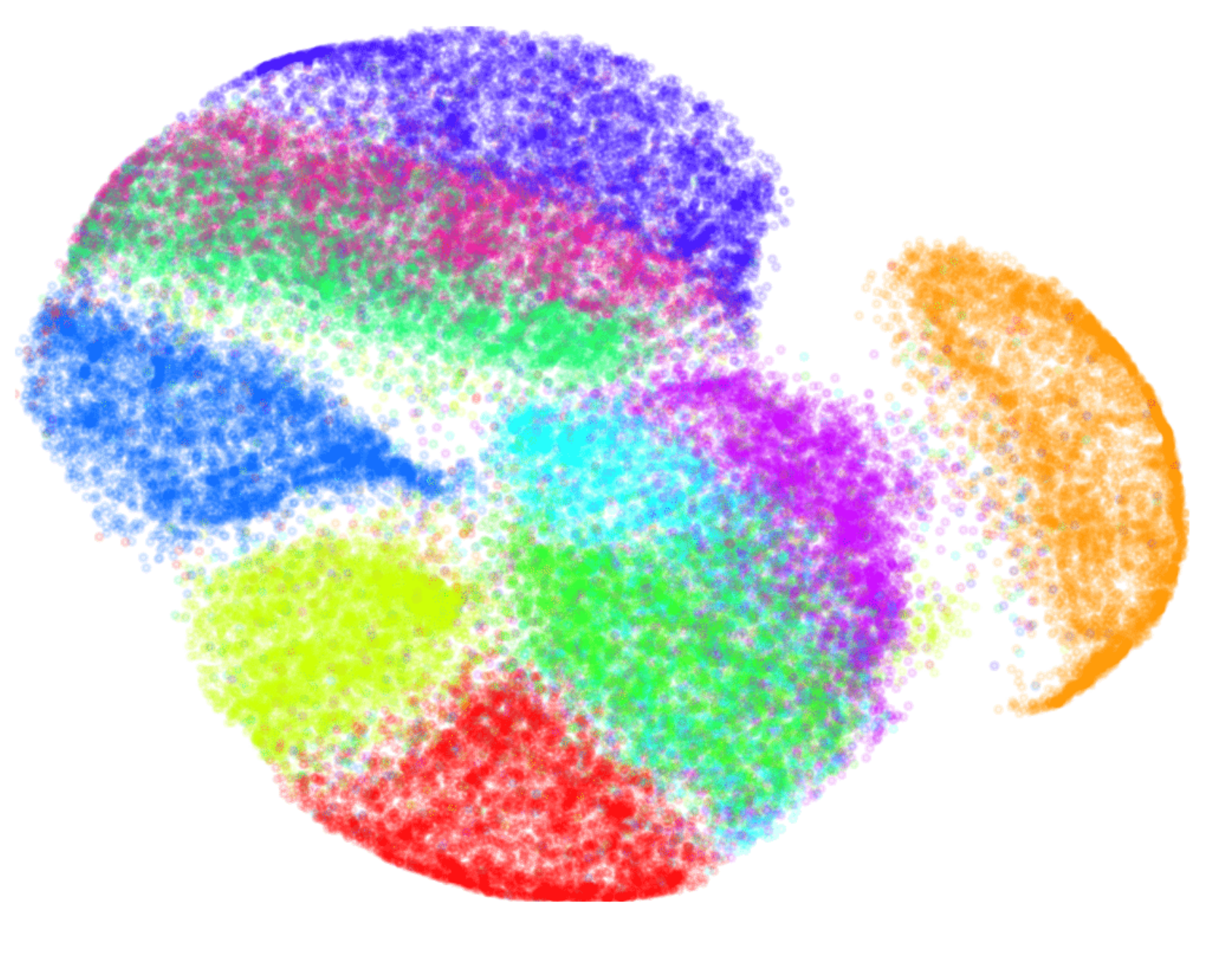}
    \hspace*{.2in}
    \includegraphics[align=c,width=0.33\textwidth]{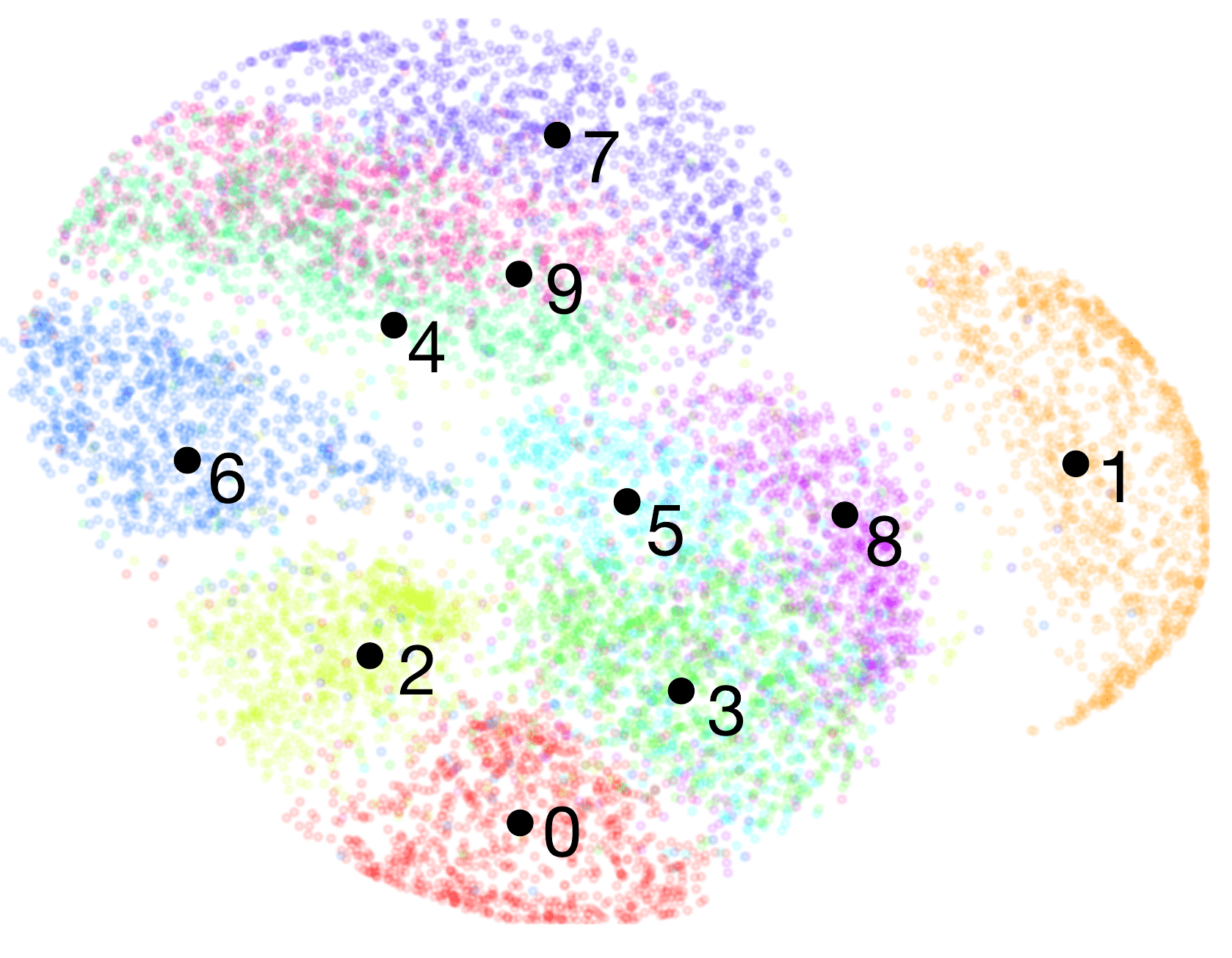}
    \hspace*{.3in}
    \includegraphics[align=c,height=1.5in]{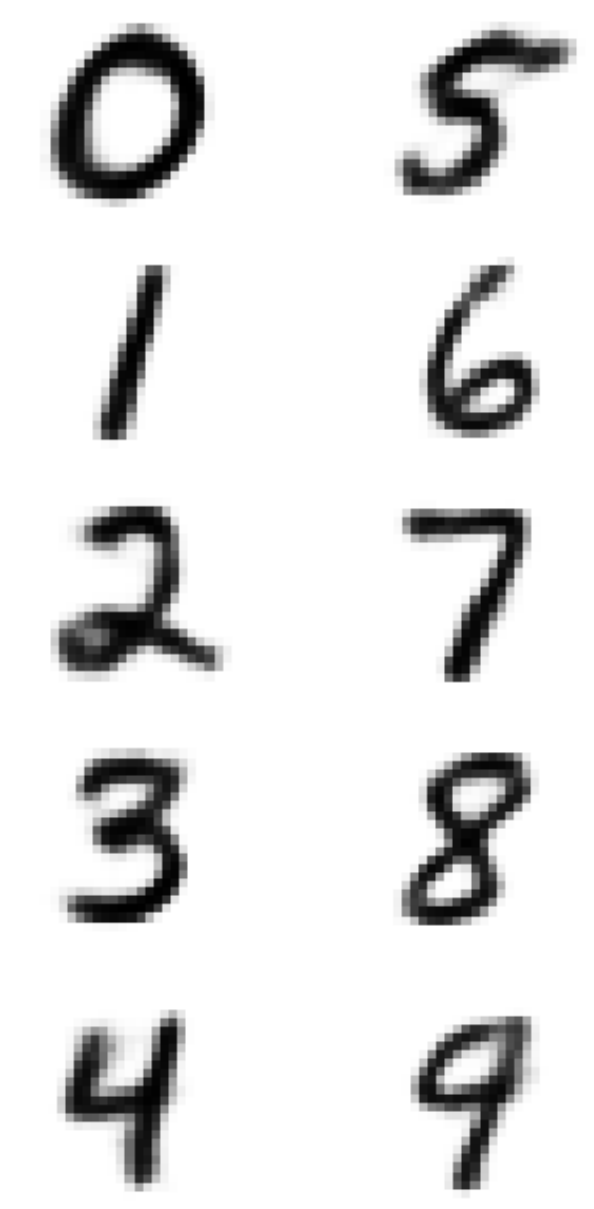}\\[2pt]
    \begin{tabular}{x{0.35\textwidth}x{0.35\textwidth}x{0.2\textwidth}}
 \textbf{(a)}~Training $X$ & \textbf{(b)}~Test $X$ & \textbf{(c)}~Samples
\end{tabular}
  \end{minipage}
  \caption{Learned $X$ space embeddings from the nonparametric VAE.
  Inferred $X$ locations for (a)~the training data and (b)~the test data with colors encoding the MNIST digit classes.
  (c)~Generated samples from the corresponding locations in (b) using a $Z$ space with 500 dimensions.
}
  \label{fig:x_generation}
\end{figure}

\begin{figure}[h]
\centering
  \begin{tabular}{m{0.5in}cccc}
  \small{VAE} &
  \begin{subfigure}{1in}\centering
  \includegraphics[height=0.6in]{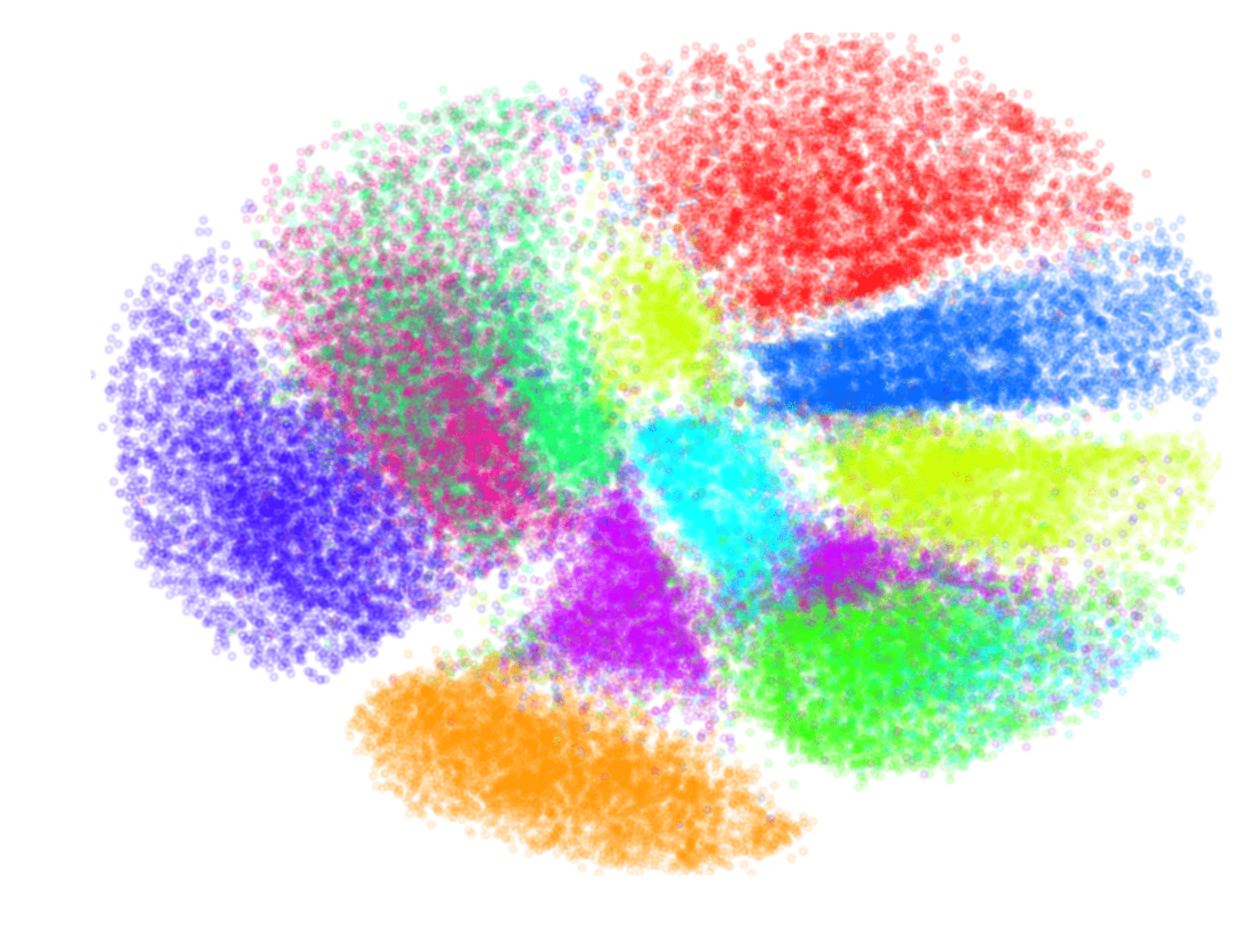}
  \end{subfigure}&
  \begin{subfigure}{1in}\centering
  \includegraphics[height=0.3in]{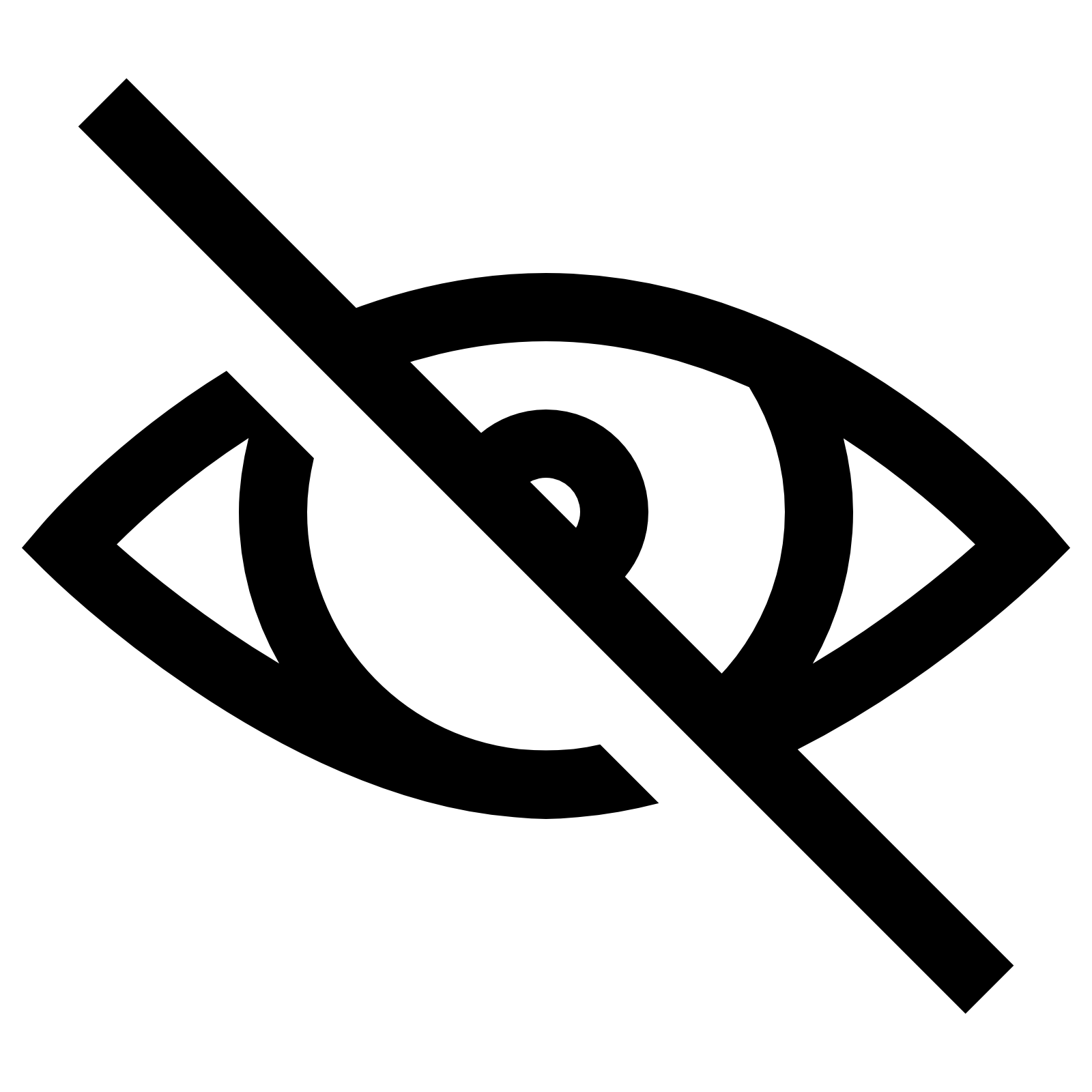}
  \end{subfigure}&
  \begin{subfigure}{1in}\centering
  \includegraphics[height=0.3in]{blind}
  \end{subfigure}&
  \begin{subfigure}{1in}\centering
  \includegraphics[height=0.3in]{blind}
  \end{subfigure}\\[5pt]
    \small{np-VAE} &
  \begin{subfigure}{1in}\centering
  \includegraphics[height=0.6in]{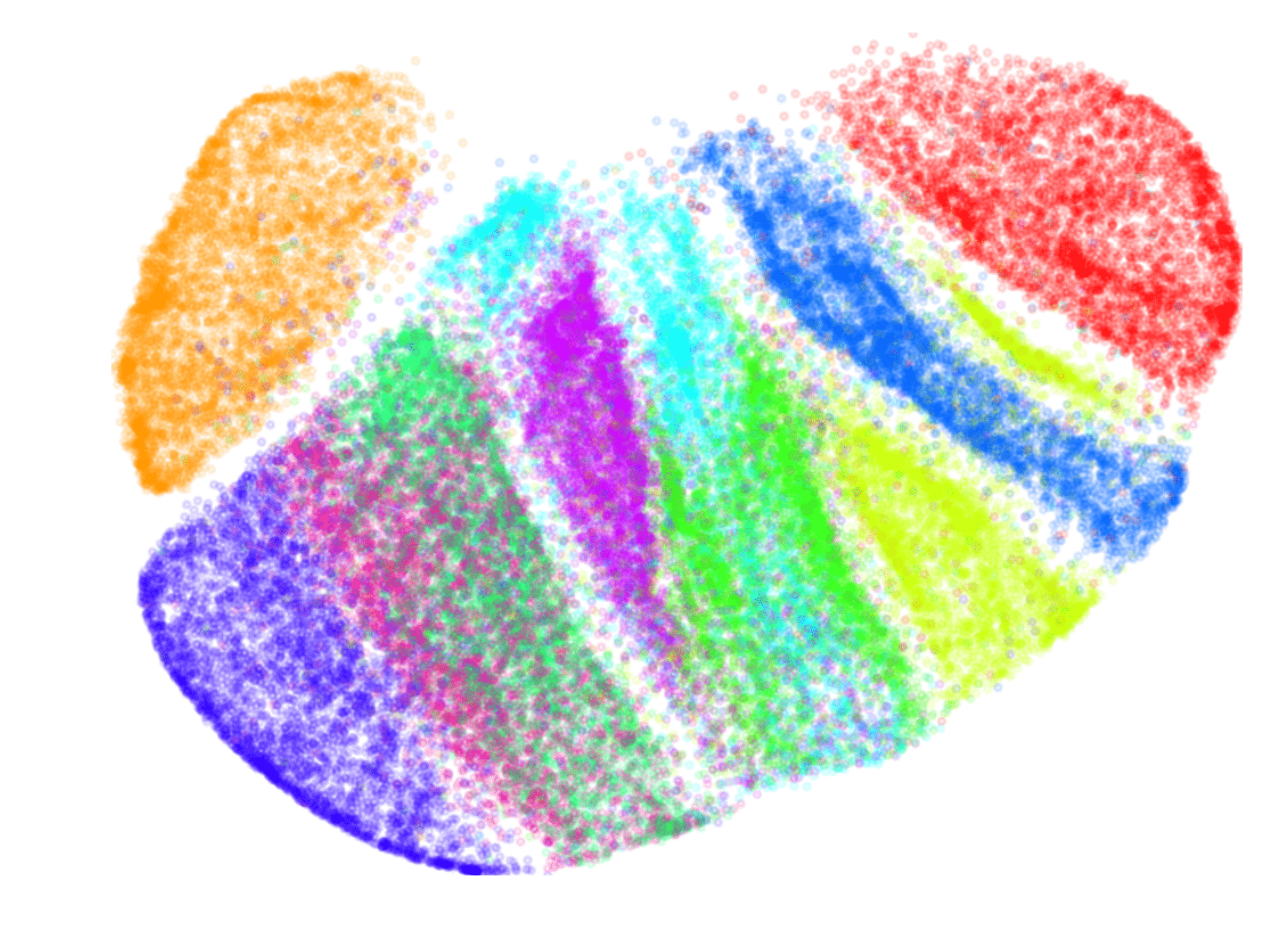}
  \end{subfigure}&
  \begin{subfigure}{1in}\centering
  \includegraphics[height=0.6in]{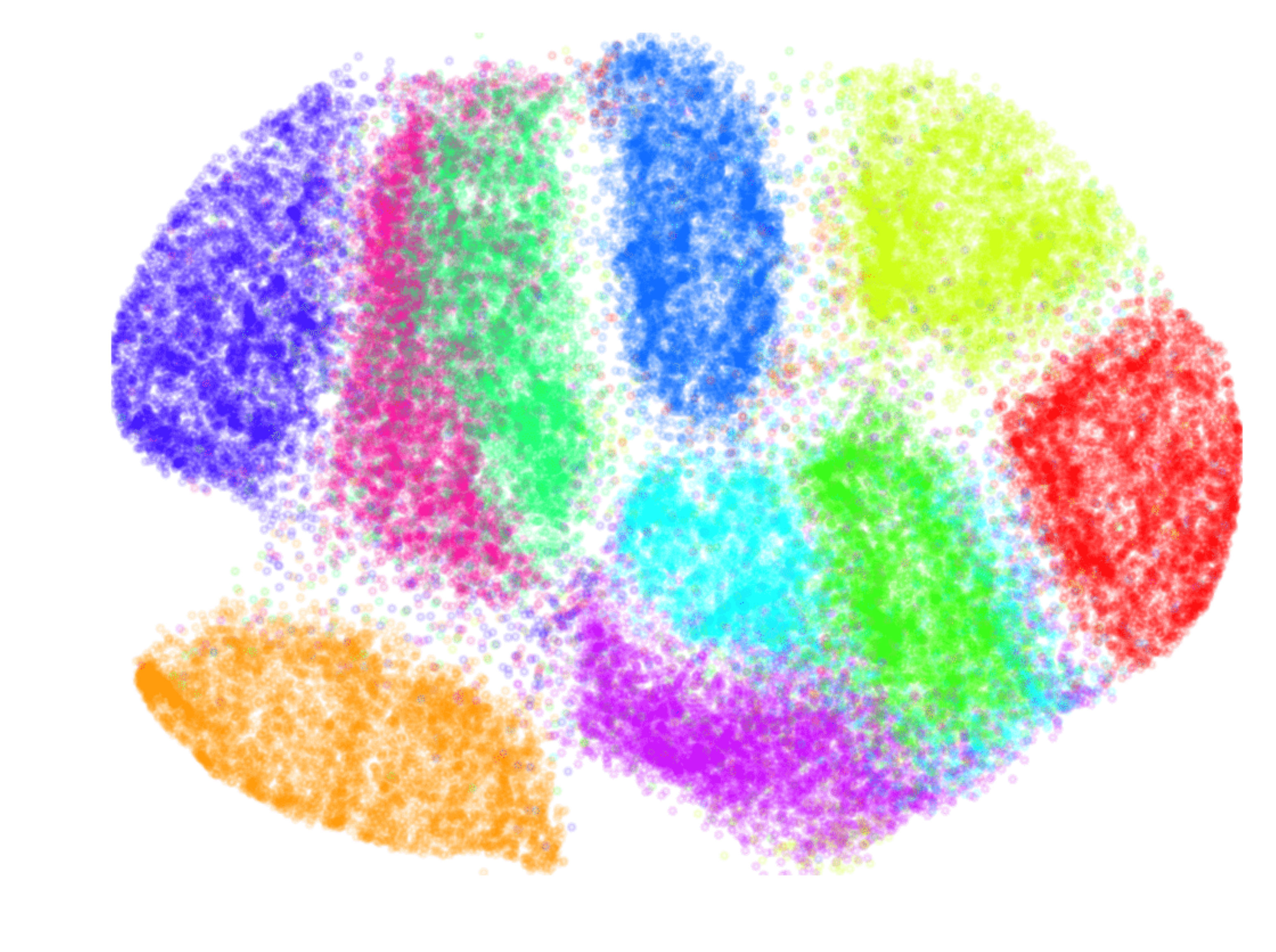}
  \end{subfigure}&
  \begin{subfigure}{1in}\centering
  \includegraphics[height=0.6in]{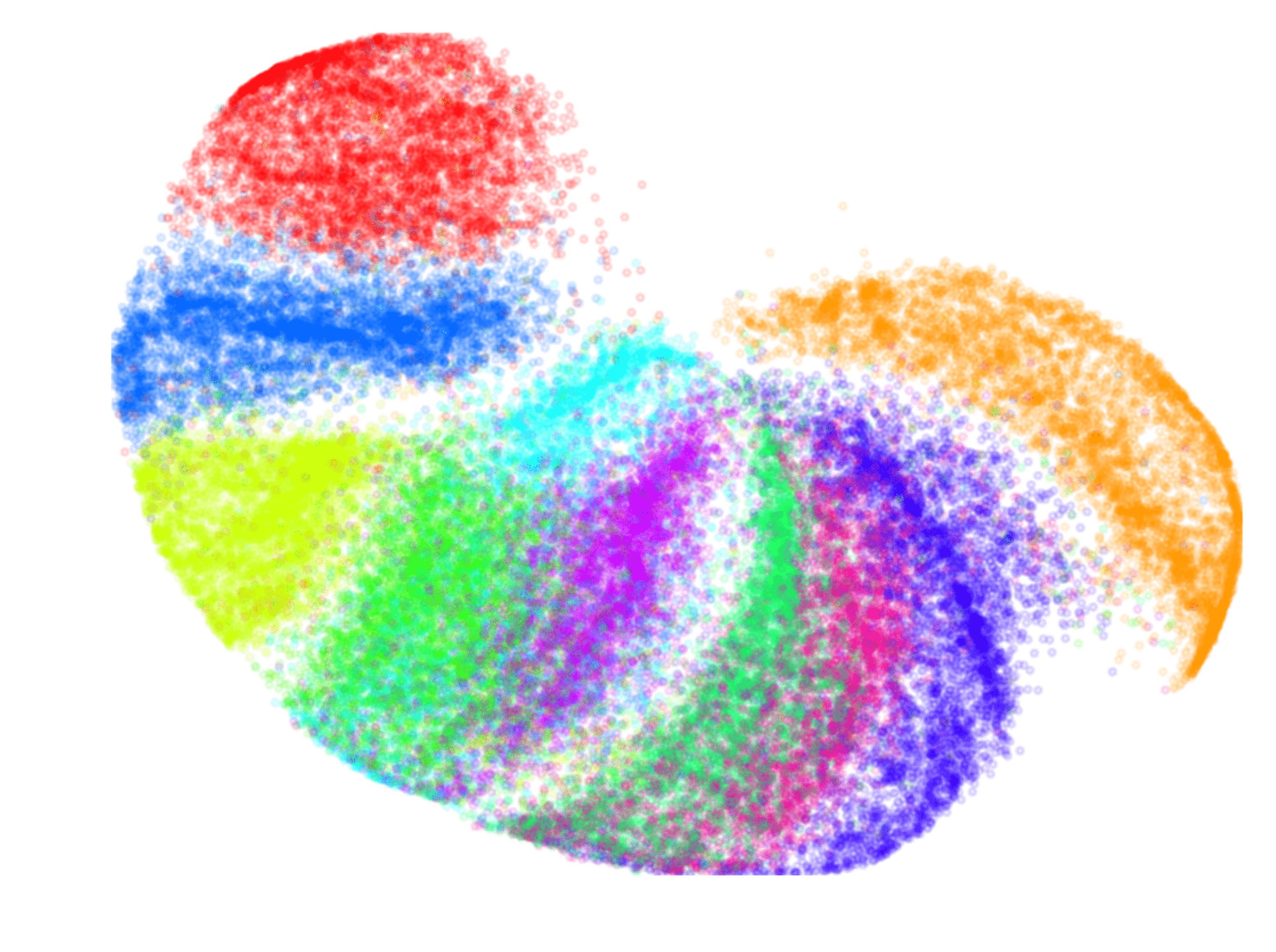}
  \end{subfigure}&
  \begin{subfigure}{1in}\centering
  \includegraphics[height=0.6in]{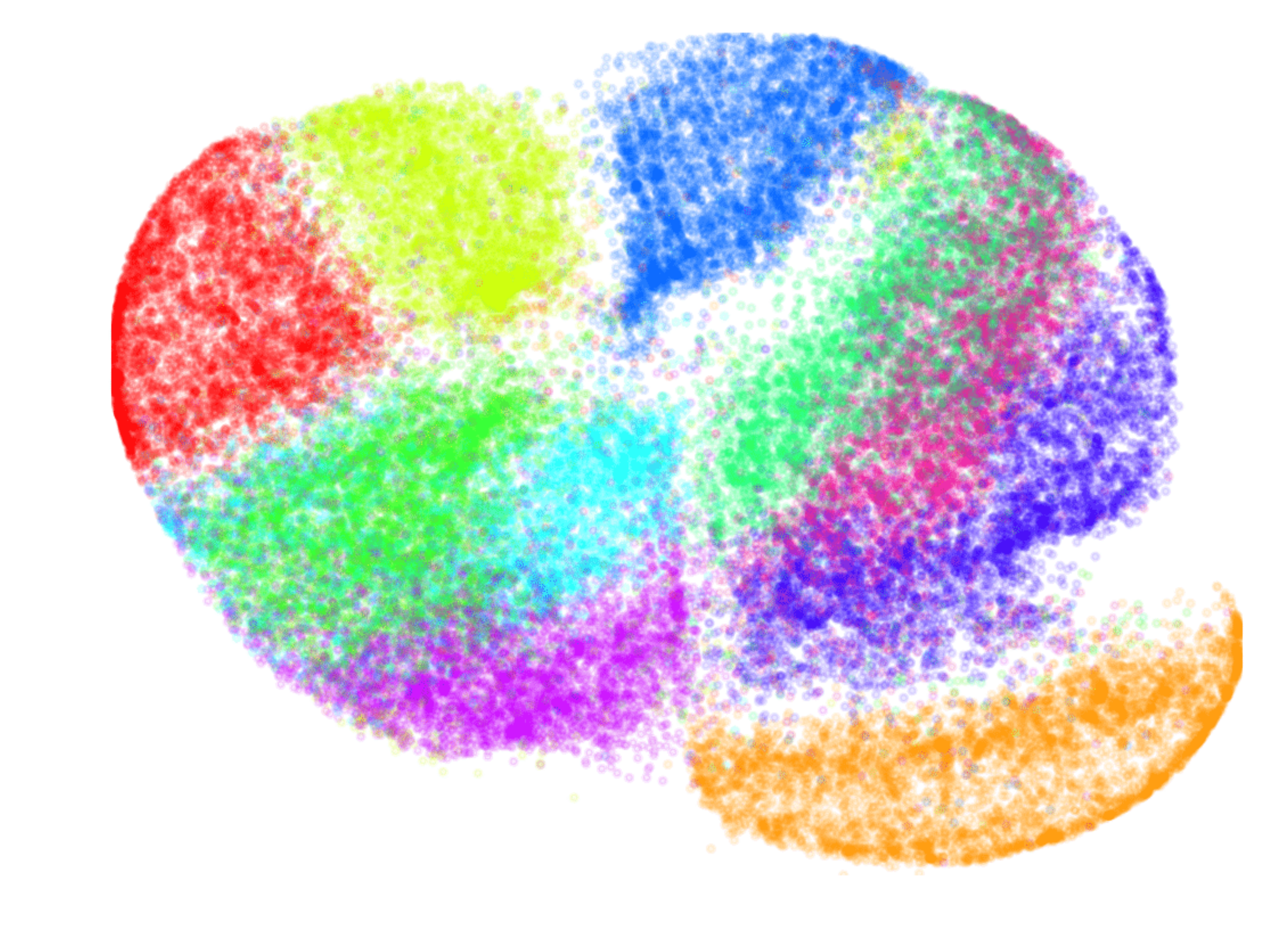}
  \end{subfigure}\\[-5pt]
  &
  \begin{subfigure}{1in}\centering
  \vspace*{.15in}
  \caption*{$\mathbf{z}_{i} \in \mathbb{R}^{2}$}
  \end{subfigure}&
  \begin{subfigure}{1in}\centering
  \vspace*{.15in}
  \caption*{$\mathbf{z}_{i} \in \mathbb{R}^{10}$}
  \end{subfigure}&
  \begin{subfigure}{1in}\centering
  \vspace*{.15in}
  \caption*{$\mathbf{z}_{i} \in \mathbb{R}^{100}$}
  \end{subfigure}&
  \begin{subfigure}{1in}\centering
  \vspace*{.15in}
  \caption*{$\mathbf{z}_{i} \in \mathbb{R}^{1000}$}
  \end{subfigure}
  \end{tabular}\\[-5pt]
\caption{
Latent space visualisation. Upper row: the $Z$ space embedding is visualised for the standard VAE where possible (we are unable to do this for high-dimensional $Z$). Bottom row: the same $Z$ space dimensionalities are used but the nonparametric VAE allows the $X$ space to be visualised and sampled (set to be $2$-dimensional). $Z$ spaces of higher dimension become impractical to visualise and interpret whereas the $X$ space provides an embedding for easy display and interpretation.
}
\label{fig:latent_spaces}
\end{figure}

\begin{figure}[ht]
    \centering
    \begin{tabular}{ll m{4in}}
        VAE & $\mathbf{z}_{i} \in \mathbb{R}^{2}$ &
        \includegraphics[align=c,width=4in]{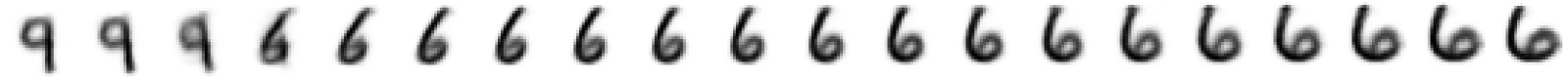} \\
         & $\mathbf{z}_{i} \in \mathbb{R}^{500}$ & 
        \includegraphics[align=c,width=4in]{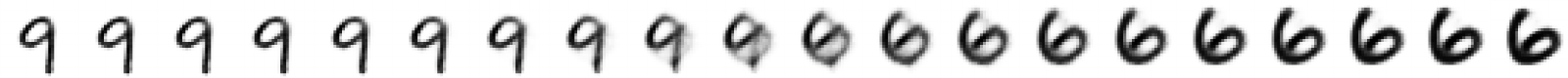} \\
        \noalign{\vskip 1.5mm}
        \hline
        \noalign{\vskip 1.5mm}
        np-VAE & $\mathbf{z}_{i} \in \mathbb{R}^{2}$ & 
        \includegraphics[align=c,width=4in]{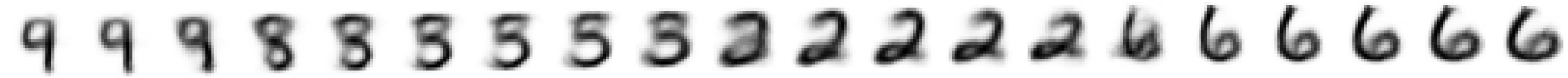}\\
        & $\mathbf{z}_{i} \in \mathbb{R}^{500}$ & 
        \includegraphics[align=c,width=4in]{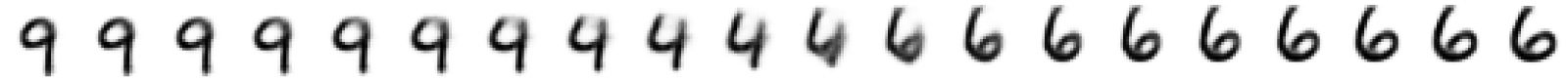} \\
    \end{tabular}
    \caption{
    Latent space interpolation.
The upper two rows show interpolants between two MNIST training examples for a standard VAE with a $Z$ latent dimensionality of 2 and 500. 
The bottom two rows show interpolants between the same training examples for our nonparametric VAE with the same respective dimensionalities for $Z$ but where the interpolation is performed in the inferred latent space $X$ of dimension 2. 
We observe that a similar reconstruction quality is obtained by corresponding $Z$-dimensionalities, however, the interpolants from the $X$ space of the nonparametric VAE are more meaningful with credible intermediate states between digits. Thus we can obtain a low dimensional latent space that provides interpretability without sacrificing reconstruction quality.
%
    }
    \label{fig:interpolations}
\end{figure}

In Fig.~\ref{fig:x_generation} we show a $2$-dimensional $X$ space corresponding to a $500$-dimensional $Z$ space with both the training and test data embedded as well as examples of data generation. 
Despite using a high capacity $Z$ space, the nonparametric VAE can still sample from a low dimension (in our example 2D) using ancestral sampling from the $X$ space. This ensures that test samples maintain a high fidelity from a space that is highly interpretable, easy to visualise and easy to sample from. 
%
%
In Fig.~\ref{fig:latent_spaces} we show corresponding $X$ space embeddings for different dimensionalities of $Z$; this demonstrates that the $X$ space maintains its virtues independent of the $Z$ dimension. Finally, in Fig.~\ref{fig:interpolations}, we show sample interpolations in the latent space for the standard VAE (directly in $Z$) and with our extension (in $X$) illustrating that not only is the reconstruction quality preserved but interpolations are more meaningful.

\section{Conclusions}

We have presented a hierarchical model for unsupervised learning and an associated efficient approximative inference scheme. The inference takes inspiration from amortised inference and use a recognition model to parameterise the approximate posterior using a deterministic relationship from the observed data. Rather than using a traditional mean-field approximation which forces the latent representation to be independent we introduce an additional latent representation that models their dependence. Our model results in a significantly lower dimensional latent representation allowing us to visualise and generate data in a intuitive manner without sacrificing the quality of the reconstruction. We have shown experimental results on how we can retain the representative power of a 500 dimensional model with just a 2 dimensional latent space.

\FloatBarrier

\bibliography{workshop}

\end{document}